\documentclass[10pt,twocolumn,letterpaper]{article}

\usepackage[pagenumbers]{cvpr} % To force page numbers, e.g. for an arXiv version
\usepackage{marvosym}

\definecolor{cvprblue}{rgb}{0.21,0.49,0.74}
\usepackage[pagebackref,breaklinks,colorlinks,allcolors=cvprblue]{hyperref}
\usepackage{multirow}
\usepackage[table]{xcolor}
\usepackage{longtable}
\usepackage{booktabs} % for better table lines

\usepackage{listings}
\usepackage{xcolor}

\lstdefinestyle{pythoncode}{
  language=Python,
  basicstyle=\ttfamily\footnotesize,
  keywordstyle=\color{blue},
  commentstyle=\color{gray},
  stringstyle=\color{teal},
  showstringspaces=false,
  breaklines=true,
  frame=single,
  framerule=0.3pt,
  rulecolor=\color{black!20},
  backgroundcolor=\color{gray!5},
  numbers=left,
  numberstyle=\tiny\color{gray},
  xleftmargin=1.5em,
  framexleftmargin=2em
}
\def\paperID{} % *** Enter the Paper ID here
\def\confName{CVPR}
\def\confYear{2026}

\makeatletter
\def\blfootnote{\xdef\@thefnmark{}\@footnotetext}
\makeatother

\title{Beyond Short-Horizon: VQ-Memory for Robust Long-Horizon Manipulation in Non-Markovian Simulation Benchmarks}

\author{
Honghui Wang$^{1\dag}$,
Zhi Jing$^{2,3}$,
Jicong Ao$^{3}$,
Shiji Song$^1$,
Xuelong Li$^3$, 
Gao Huang$^{1}$\textsuperscript{\Letter},
Chenji Bai$^{3}$\textsuperscript{\Letter}\\
$^1$Tsinghua University, $^2$Fudan University\\
$^3$Institute of Artificial Intelligence (TeleAI), China Telecom\\
}

\begin{document}
\maketitle
\begin{abstract}
The high cost of collecting real-robot data has made robotic simulation a scalable platform for both evaluation and data generation. Yet, most existing benchmarks concentrate on simple manipulation tasks like pick-and-place, failing to capture the non-Markovian characteristics of real-world tasks and the complexity of articulated object interactions.
To address this limitation, we present RuleSafe, a new articulated manipulation benchmark built upon a scalable, LLM-aided simulation framework. RuleSafe features safes with diverse unlocking mechanisms—such as key, password, and logic locks—that require distinct multi-stage reasoning and manipulation strategies. These LLM-generated rules yield non-Markovian, long-horizon tasks that demand temporal modeling and memory-based reasoning.
We further propose VQ-Memory, a compact and structured temporal representation that leverages vector-quantized variational autoencoders (VQ-VAEs) to encode past proprioceptive states into discrete latent tokens. This representation effectively filters low-level noise while preserving high-level task-phase context, providing lightweight yet robust temporal cues that are compatible with existing Vision-Language-Action models (VLA). Extensive experiments on state-of-the-art VLA models and diffusion policies demonstrate that VQ-Memory consistently improves long-horizon planning, enhances generalization to unseen configurations, and achieves more efficient manipulation with reduced computational cost.
Project page is \url{https://vqmemory.github.io}.

\end{abstract}   
\blfootnote{\noindent $^{\dag}$This work is done when Honghui Wang is intern at TeleAI. \textsuperscript{\Letter}Corresponding authors: Chenjia Bai (baicj@chinatelecom.cn), Gao Huang (gaohuang@tsinghua.edu.cn)}

\section{Introduction}
\label{sec:intro}

Robotic simulation has recently achieved remarkable progress, establishing itself as a scalable platform for both evaluation and data generation~\cite{wanggensim,nasiriany2024robocasa,mu2024robotwin}. This advancement has been driven by innovations such as leveraging large language models (LLMs) for automated data generation~\cite{hua2025gensim2}, employing data augmentation to narrow the sim-to-real gaps~\cite{chen2025robotwin}, and incorporating complex embodiments like dual-arm and humanoid robots~\cite{jing2025humanoidgen}. 
However, existing simulation benchmarks still predominantly focus on short-horizon tasks and simple manipulations (e.g., pick-and-place), resulting in limited diversity of object interactions. Critically, articulated objects—such as doors, drawers, and cabinets—are ubiquitous in real-world environments and introduce substantial manipulation complexity, yet they remain underexplored in current benchmarks.

Previous articulated manipulation benchmarks~\cite{xiang2020sapien,geng2023gapartnet,wang2025articubot} have concentrated on evaluating a model's ability to infer affordances and joint constraints of unseen objects from visual observations. 
Crucially, these studies rarely account for inter-joint dependencies or the interplay among multiple kinematic components of the same object. This omission significantly limits the diversity and complexity of possible actions. For instance, while manipulating a door often only involves simple opening or closing motions, introducing a door lock immediately expands the action space, demanding sequential reasoning over both unlocking and subsequent opening. 
Although some efforts have incorporated such mechanisms (e.g., door locks to control object states~\cite{li2024unidoormanip,wangadamanip}), these approaches rely on manually scripted rules for trajectory collection. This manual process inherently constrains their extensibility and scalability compared to benchmarks utilizing LLM-aided generation, and ultimately results in the predominance of short-horizon tasks.

In this paper, we introduce \textbf{RuleSafe}, a new articulated manipulation benchmark built upon a scalable, LLM-aided simulation framework. RuleSafe features a collection of safes equipped with diverse unlocking mechanisms, such as key locks, password locks, and logic locks, each demanding distinct reasoning and manipulation strategies. 
% These mechanisms are governed by rules systematically designed based on 1) the part phases of the articulated components and 2) the defined task phase that track the progress of the multi-stage process. 
These mechanisms are governed by rules systematically designed based on (1) the \textit{part-phases} of articulated components, defined as discrete states (e.g., open or closed) derived from their kinematic states, such as rotation angles or translation distances, and (2) the \textit{task-phase}, which tracks the progress of the multi-stage process.
By imposing these multi-step dependencies, RuleSafe significantly expands the planning space and enables the robust evaluation of policies’ ability to plan and execute long-horizon interactions. 
To ensure scalability and minimize manual effort, these rules are predominantly generated by LLMs from only a few examples. 

RuleSafe distinguishes itself from prior benchmarks as a challenge platform demanding temporal modeling. Specifically, the non-Markovian nature of the tasks arises because the current task stage and the joint states of the safe components cannot be inferred from a single visual observation. 
Consequently, successful manipulation requires policies to reason over temporal sequences and maintain memory. Through extensive evaluations of state-of-the-art Vision-Language-Action (VLA) models~\cite{black2024pi0,li2024cogact,liurdt} and diffusion policies~\cite{ze20243ddiffusion}, we observe a consistent failure pattern: policies relying solely on the current visual frame struggle to differentiate visually similar yet semantically distinct manipulation stages. While incorporating visual history mitigates this, it introduces significant computational overhead and scalability concerns. Conversely, using a history of raw robot \textbf{joint states} (i.e., the robot’s proprioceptive states representing each joint configuration) provides a lightweight temporal cue but is susceptible to low-level noise and prone to overfitting specific trajectories.

To address these limitations, we propose VQ-Memory, a compact and structured temporal representation that leverages vector-quantized variational autoencoders (VQ-VAEs)~\cite{van2017neural,zeghidour2021soundstream} to encode raw joint-state sequences into discrete latent tokens. These discrete memory representations effectively filter out irrelevant low-level variations while preserving high-level phase context, enabling the policy to maintain robust stage awareness. 
Notably, VQ-Memory is model-agnostic and can be incorporated into a variety of VLA models and diffusion policies, demonstrating its generality across different architectures.
Empirically, VQ-Memory enhances long-horizon planning, improves generalization to unseen configurations, and achieves efficient manipulation with reduced computational cost. 

In summary, the key contributions of this paper are:
\begin{itemize}
    \item We introduce \textbf{RuleSafe}, a novel, LLM-aided articulated manipulation benchmark featuring non-Markovian tasks
    that demand long-horizon planning.
    \item We propose \textbf{VQ-Memory}, a compact and structured temporal memory using VQ-VAEs to robustly encode high-level task phase context from noisy joint-state histories.
    \item We demonstrate that VQ-Memory is a model-agnostic module that significantly enhances the long-horizon planning and generalization capabilities of various VLA.
\end{itemize}

\section{Related Works}
\label{sec:related}

\subsection{Robotic Simulation Benchmarks}

Recent advances in robotic simulation have established it as a powerful and scalable environment for systematic evaluation and large-scale data generation~\cite{wanggensim,nasiriany2024robocasa,mu2024robotwin,wangrobogen,huang2025rekep}, largely fueled by LLMs. 
Benchmarks such as GenSim~\cite{wanggensim} and RoboCasa~\cite{nasiriany2024robocasa} enrich the task diversity by leveraging LLMs to automatically generate task descriptions. 
RoboTwin~\cite{mu2024robotwin,chen2025robotwin} broadens the scope to dual-arm manipulation and improves sim-to-real transfer through domain randomization, while HumanoidGen~\cite{jing2025humanoidgen} further advances this paradigm by introducing complex embodiments such as humanoid robots.
Despite these efforts, most existing benchmarks still focus on short-horizon or low-level tasks (e.g., pick-and-place).
GenSim2~\cite{hua2025gensim2} incorporates articulated objects to increase task variety. However, the interactions remain constrained to simple motion primitives, mostly achieved through arm translation, and lack precise part-specific interactions (e.g., rotating knobs).
Building upon the HumanoidGen framework, our work focuses on fine-grained manipulation of articulated objects and leverages multi-joint dependencies to design multi-stage tasks.

\subsection{Articulated Object Manipulation}
Several benchmarks have been proposed for articulated object manipulation~\cite{xiang2020sapien,geng2023gapartnet,li2024unidoormanip,wang2025articubot}. PartNet-Mobility~\cite{xiang2020sapien} provides multiple categories of articulated objects derived from PartNet~\cite{mo2019partnet}. GAPartNet~\cite{geng2023gapartnet} emphasizes part-level details, offering rich, fine-grained annotations for individual object components. UniDoorManip~\cite{li2024unidoormanip} targets door environments, encompassing diverse door bodies and handle types to support more varied interaction scenarios.
These benchmarks are predominantly designed to assess a model's capacity for cross-object generalization, specifically its proficiency in inferring visual affordances and joint constraints for novel articulated objects. A significant limitation, however, is their confinement to single-joint motions, thereby constraining the action diversity within the evaluated tasks.
Although AdaManip~\cite{wangadamanip} increases task complexity by incorporating multi-joint dependencies like lock mechanisms, its reliance on manually scripted trajectories limits scalability and hinders the creation of long-horizon tasks.
We tackle this limitation by creating diverse unlocking rules to broaden the planning space and employing an LLM-aided framework for scalable, long-horizon data generation.

\subsection{Temporal Modeling in Robotic Policies}

Manipulating articulated objects with multi-step dependencies requires temporal reasoning, as the current observation alone is insufficient to gauge task progress. 
Such long-horizon dependencies pose a particular challenge for Vision-Language-Action (VLA) models~\cite{brohan2023rt,zitkovich2023rt,kimopenvla,livision}.
% Recent Vision-Language-Action (VLA) models~\cite{brohan2023rt,zitkovich2023rt,livision} address this challenge by integrating multimodal inputs for long-horizon reasoning.
Recent works like Octo~\cite{mees2024octo}, RDT~\cite{liurdt}, and RoboVLM~\cite{li2024towards} incorporate historical visual observations to enhance temporal reasoning. MindExplore~\cite{li2025towards} further leverages memory for feedback and error correction. However, using raw visual observations as memory entails substantial computational costs. MemoryVLA~\cite{shi2025memoryvla} addresses this by compressing historical information into a fixed-size memory bank. In contrast, TraceVLA~\cite{zhengtracevla} represents past states as trajectories overlaid on the current frame, which is efficient but limited in long-context capture.
Alternatively, AdaManip~\cite{wangadamanip} proposes augmenting diffusion policies with historical joint states. 
Although joint-state sequences provide a lightweight form of temporal context, we find they are noise-sensitive and can cause overfitting to training sequences. 
Our proposed VQ-Memory addresses this by discretizing past joint states into structured tokens via a VQ-VAE~\cite{van2017neural}, 
providing a compact, robust, and model-agnostic temporal memory.
\section{RuleSafe}

Existing simulation environments for articulated manipulation are typically hindered by two major limitations: their lack of multi-step, interdependent action sequences needed to capture realistic processes, and their heavy reliance on manual task logic, which severely restricts diversity and scalability.
To address these challenges, we introduce RuleSafe, a scalable and systematically structured benchmark for long-horizon manipulation. 
RuleSafe provides a diverse suite of articulated locking mechanisms (\cref{sec:lock_gen}), including key, password, and logic locks, and employs an LLM-aided data generation pipeline (\cref{sec:demon_gen}) to ensure both scalability and diversity. See Appendix~\ref{sec:rule_and_obj} for more details.

\subsection{Locking Mechanism Generation}
\label{sec:lock_gen}

\begin{figure}[t]
  \centering
   \includegraphics[width=0.99\linewidth]{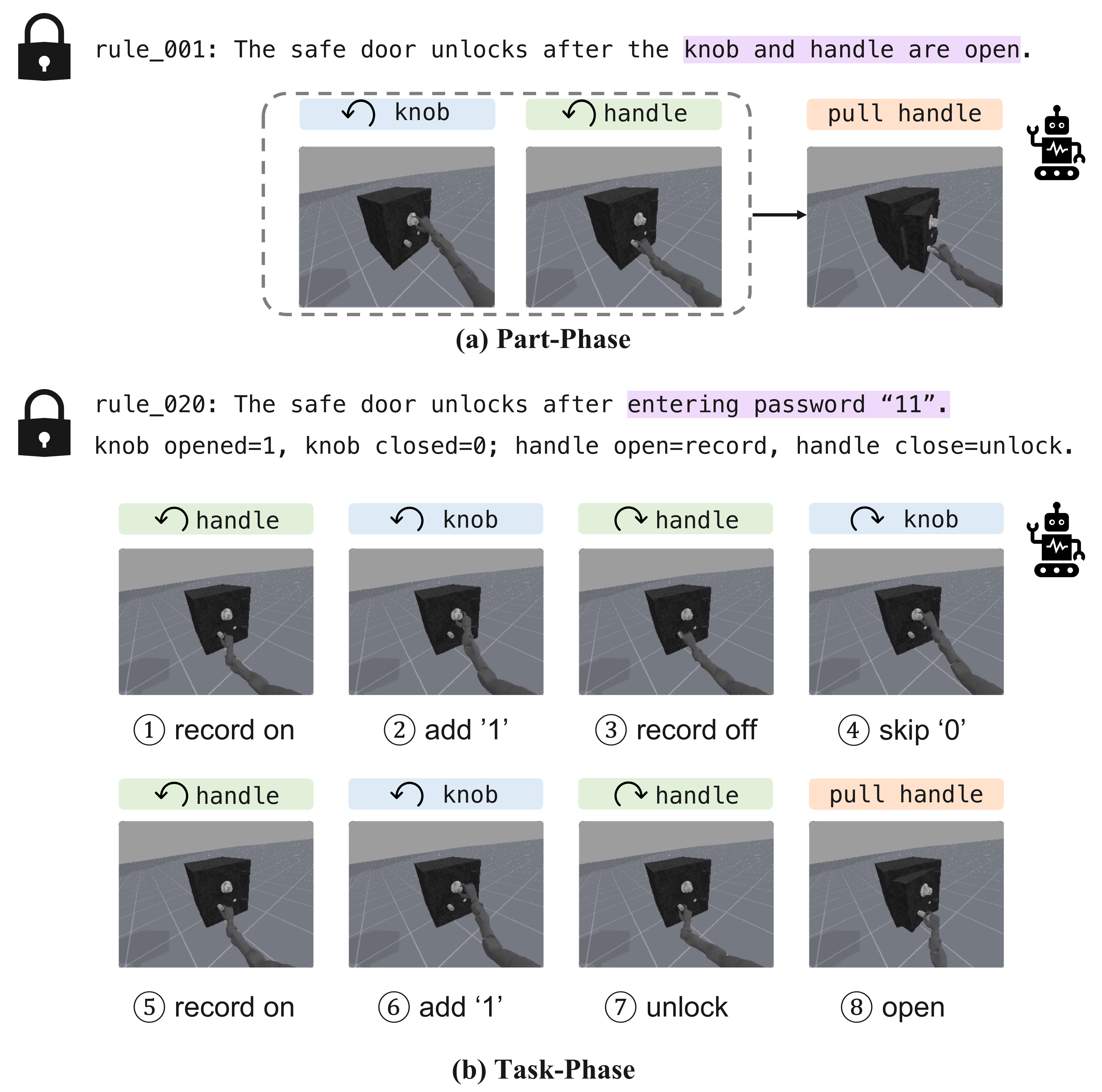}
   \vspace{-0.8em}
   \caption{Locking rules from the RuleSafe Benchmark, structured by (a) part phase (knob/handle) and (b) task phase (password cache). Because these governing variables are \textit{not} directly observable from visual input, the tasks exhibit a non-Markovian property.}
   \label{fig:two_rules}
   \vspace{-1.5em}
\end{figure}

The design of locking mechanisms in RuleSafe is driven by the need to support multi-step, interdependent manipulation tasks. To systematically construct such mechanisms, we base the designs on two types of latent variables: the \textit{part-phase} of articulated objects and the \textit{task-phase} that represents execution progress. These latent variables provide the underlying structure to define complex dependencies between object components and multi-stage actions, enabling diverse and challenging manipulation tasks.

\paragraph{Part-Phase}

The part-phase captures the configuration of an object’s articulated components as discrete states (e.g., open or closed) derived from their underlying kinematic states, such as the rotation of a knob or the translation of a drawer.
By tracking the states of individual parts, we can define sequential dependencies between actions as shown in~\cref{fig:two_rules}(a). For instance, a simple key-lock can be represented as a specific part reaching a target angle. More complex structures, such as a chained key-lock (e.g., knob open $\rightarrow$ handle open $\rightarrow$ door open), require actions to be executed in a precise order across multiple joints. 

\paragraph{Task-Phase} 
The task-phase provides a higher-level abstraction that tracks progress across multi-step tasks. Unlike part-phase, it does not directly correspond to individual joint configurations, but instead encodes whether specific stages of a task have been completed. Task-phase enables the creation of more flexible locking mechanisms, such as password locks or logic locks, where the unlocking condition may depend on combinations of multiple actions or input sequences, as shown in~\cref{fig:two_rules}(b). 
% For example, a password lock may require a user to rotate two dials to specific values in order to unlock a safe.
For example, a password lock may require the agent to manipulate several object parts following specific rules in order to enter the password.
A logic lock, on the other hand, could enforce a relational constraint — for instance, the total number of knob and handle rotations must equal a target value.

\paragraph{LLM-Aided Lock Design} 
To efficiently generate a wide variety of locking mechanisms, we leverage LLMs to design locks centered around both part-phase and task-phase specifications.
We first manually design a set of reference examples, which serve as guidance for the LLM. Given these examples, the LLM outputs two components for each lock: (1) a description of the rules, which can later be used as a prompt for manipulation policies, and (2) an executable program that determines whether the unlocking conditions are met (see Appendix~\cref{lst:check_lock}). We also provide a template for the unlocking judgment program, which uses updates in the part-phase to check whether the task-phase should be updated, and ultimately verifies whether the task has been successfully completed. This approach significantly reduces manual effort while maintaining scalability and diversity in the types of locks available within the benchmark.

\subsection{Demonstration Generation}
\label{sec:demon_gen}

\begin{figure*}[t]
  \centering
   \includegraphics[width=0.99\linewidth]{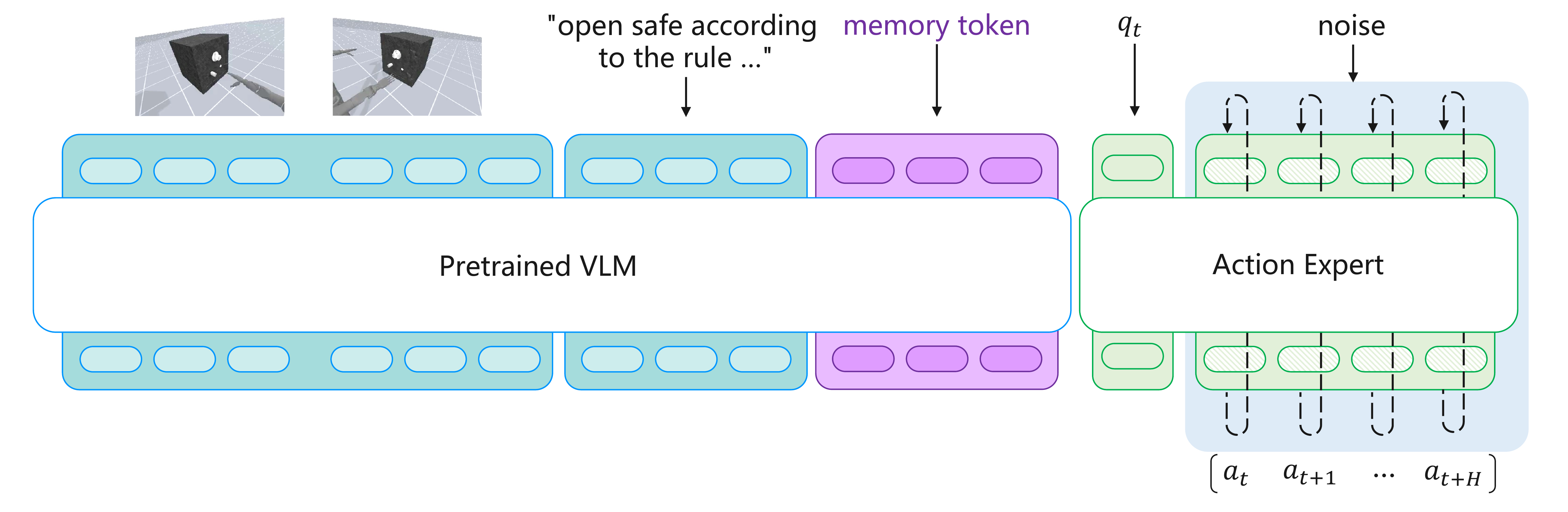}
   \vspace{-1.5em}
   \caption{Overview of the memory-augmented policy formulation. The pretrained Vision-Language Model encodes current observations and task instructions, while the memory tokens $\boldsymbol{m}_t$ provide additional historical context from previous timesteps. The combined representation is fed into the Action Expert, which predicts the future action sequence conditioned on the state $q_t$ and injected noise. }
   \label{fig:framework}
   \vspace{-1.5em}
\end{figure*}

Demonstrations in RuleSafe are generated by extending the capabilities of the HumanoidGen~\cite{jing2025humanoidgen} framework on articulated manipulation. The process involves spatial annotation of atomic operations and assets, as well as task decomposition for long-horizon manipulation.

\paragraph{Task Decomposition} 
For each lock rule, an LLM-based planner decomposes the manipulation task into a sequence of action operations $S=\{S_1, S_2, \ldots, S_n\}$, where each step $S_i$ is either a hand operation $A_i$ or an arm movement $M_i$. 
This decomposition is performed via code generation~\cite{liang2022code,mu2024robocodex} (see Appendix Listing~\ref{lst:llm-decomp}).
A hand operation $A_i$ is drawn from a library of atomic actions $A^{\text{hand}}=\{A^{\text{pinch}}, A^{\text{grasp}}, \ldots, A^{\text{rotate}}\}$. An arm movement $M_i$ is formulated as a constrained optimization problem, with the constraints generated by the LLM to specify the target end-effector pose. During simulation, the actual goal pose of the end-effector is obtained by solving this optimization problem and is allowed to deviate from the target pose within a tolerance range for improved flexibility. Intermediate poses between the current and goal configurations are then generated using an OMPL-based motion planner~\cite{sucan2012the-open-motion-planning-library}.

\paragraph{Spatial Annotation} 
Spatial annotation specifies the interaction of dexterous hands with articulated assets by defining the geometric constraints on arm movements. Specifically, each arm movement $M_i$ is formulated with two types of spatial constraints: point constraints and axis constraints. To enable these constraints, both the hand’s atomic operations and the articulated assets are spatially annotated with corresponding key points and axes, as shown in Appendix~\cref{fig:annotation}.
For each hand atomic operation, we annotate key points and axes that represent the contact locations and motion directions of the fingers. Similarly, for each manipulable joint of an articulated asset, we annotate key points and axes that indicate its operational interfaces.
The spatial relationship is established by aligning these annotated components. Key points enforce positional correspondence via point constraints, while axes regulate orientation (such as parallel or perpendicular alignment) via axis constraints. Collectively, these annotations ensure that the optimized arm movement is physically plausible and geometrically consistent across diverse tasks.

\paragraph{Benchmark Details} 
RuleSafe uses the SAPIEN simulation engine~\cite{xiang2020sapien} and comprises 20 lock rules executed with Inspire hands on the arms of the Unitree H1-2 humanoid robot, which provides 7 DoFs per arm and 6 DoFs per hand (13-dimensional action space). We collect RGB images from first- and third-person views, along with point clouds.

\section{Methods}

\begin{figure*}[t]
  \centering
   \includegraphics[width=0.99\linewidth]{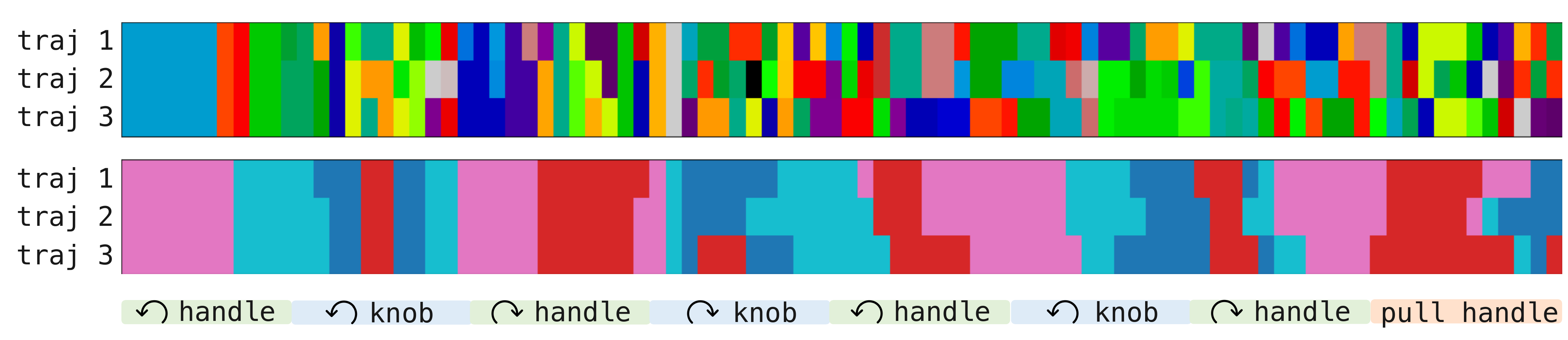}
   \vspace{-1em}
   \caption{Visualization of discrete tokens of joint trajectories with (\textit{Bottom}: vocabulary size = 4) and without clustering (\textit{Top}: vocabulary size = 256). Each column represents a time step, and each row corresponds to one of three different trajectories, where different colors indicate different tokens. Without clustering, high-level temporal regularities are obscured by fine-grained variations, whereas clustering emphasizes shared semantic patterns across trajectories, enabling clearer identification of task execution stages. }
   \label{fig:clustering}
   \vspace{-1em}
\end{figure*}

\subsection{Preliminaries}

\paragraph{Problem Formulation}
We consider a sequential decision-making problem in which an embodied agent interacts with its environment through visual observations and task-specific or language-based instructions.
The objective is to learn a policy $\pi(\boldsymbol{a}_t | \boldsymbol{o}_t)$ that predicts a chunk of future actions~\cite{zhao2023learning} $\boldsymbol{a}_t = \{a_t, a_{t+1}, \dots, a_{t+H-1}\}$ conditioned on the current observation $\boldsymbol{o}_t$.
Each observation $\boldsymbol{o}_t = \{I_t^1, I_t^2, l_t, \boldsymbol{q}_t\}$ comprises two RGB images $I_t^1$ and $I_t^2$ captured from different viewpoints, a language instruction $l_t$, and the robot’s proprioceptive state $\boldsymbol{q}_t$.

\paragraph{Temporal Modeling with Memory Tokens}
The formulation of $\pi(\boldsymbol{a}_t|\boldsymbol{o}_t)$ inherently assumes a Markovian decision process, where the current observation $\boldsymbol{o}_t$ is sufficient to determine the optimal action.
However, this assumption often fails in real-world embodied tasks that exhibit partial observability (part-phase) and long-horizon temporal dependencies (task-phase) as characterized in our benchmark.
A natural solution is to extend the policy to incorporate additional historical context, resulting in a memory-augmented formulation $\pi(\boldsymbol{a}_t|\boldsymbol{o}_t, \boldsymbol{m}_t)$, where $\boldsymbol{m}_t$ denotes a set of \textit{memory tokens} that encode information from previous timesteps.

Previous studies have explored different forms of memory representations.
Some approaches~\cite{li2024towards,mees2024octo} leverage past visual frames as memory inputs, which integrate naturally with the VLM framework but require maintaining a large number of tokens, leading to substantial computational overhead.
% Alternatively, other work~\cite{wangadamanip} represent memory using past actions, which are low-dimensional and thus more compact.
In contrast, representing memory through past joint states~\cite{wangadamanip} is far more efficient, as the robot's joint states are low-dimensional.
Motivated by this efficiency advantage, we focus exclusively on joint-state memory, which offers a lightweight yet expressive way to retain temporal context without inflating token length or computation cost.
However, as joint-state sequences are continuous signals, they are inherently susceptible to noise and can easily cause overfitting, especially in long-horizon tasks.

\subsection{VQ-Memory}
To address these limitations, we propose VQ-Memory, which learns a compact and semantically structured memory representation from joint-state sequences by combining discrete tokenization with a post-hoc clustering mechanism, enabling efficient temporal reasoning while mitigating overfitting from noisy continuous joint-state sequences.

\paragraph{Discrete Joint-State from VQVAE}
We employ a Vector-Quantized Variational Autoencoder (VQVAE)~\cite{van2017neural} to learn a discrete and structured representation of the joint-state sequence. 
Given a continuous joint-state trajectory $\boldsymbol{Q}_t = \{\boldsymbol{q}_{t-W+1}, \dots, \boldsymbol{q}_{t-1},\boldsymbol{q}_{t}\}$, the encoder maps it into a latent embedding $z_t=\mathcal{E}(\boldsymbol{Q}_t)$, which is then quantized to the nearest codebook entry $\boldsymbol{e}^k_t$ in a learned dictionary $E = \{\boldsymbol{e}^1, \boldsymbol{e}^2, \dots, \boldsymbol{e}^K\}$. 
The decoder reconstructs the original joint states $\boldsymbol{\hat{Q}}_t=\mathcal{D}(\boldsymbol{e}^k_t)$ from the quantized embedding, ensuring the learned codebook captures the essential motion primitives underlying the task. This framework is trained to minimize the reconstruction loss $L_{\mathrm{recon}}$ and the commitment loss $L_{\mathrm{commit}}$, formulated as
\begin{equation*}
    \begin{aligned}
     L &= L_{\mathrm{recon}} + \lambda L_{\mathrm{commit}} \\
     &= \|\boldsymbol{Q}_t - \boldsymbol{\hat{Q}}_t \|_2^2 + \lambda (\|z_t - \mathrm{sg}(\boldsymbol{e}^k_t) \|_2^2+\|\mathrm{sg}(z_t) - \boldsymbol{e}^k_t \|_2^2),
\end{aligned}
\end{equation*}
where $\mathrm{sg}(\cdot)$ denotes the stop-gradient operation, and the balancing coefficient $\lambda$ is set to 4.

While the previous study~\cite{wang25vqvla} employed VQVAE as a tokenizer for action sequences, its primary goal was to use discrete tokens in the policy’s output space to improve optimization efficiency and accurately reconstruct the continuous action. To maintain reconstruction fidelity, it typically adopts a small window size $W$ with an equal stride, resulting in a low token compression ratio (e.g., 5:4, where five action steps are converted into four tokens).
In contrast, our method leverages these discrete tokens solely as auxiliary inputs to provide temporal context, without requiring exact reconstruction of the raw signals. Therefore, we adopt a larger window and stride to reduce the token count and enhance computational efficiency. In practice, we set the window size to 50 and the stride to 20, achieving approximately a 20× compression ratio.

\paragraph{Compact Memory via Clustering}
While the VQVAE transforms continuous joint trajectories into a finite discrete space, it cannot fully resolve the overfitting issue. The learned codebook often contains redundant entries, leading to fine-grained variations in discrete token sequences that obscure high-level temporal regularities, as shown in~\cref{fig:clustering}. To address this, we apply a simple yet effective clustering step to the codebook after the training of VQVAE.
Specifically, we perform $K$-means clustering over the learned codebook $\{\boldsymbol{e}_k\}_{k=1}^{K}$ to merge redundant codes and form a more compact vocabulary $\{\boldsymbol{c}_j\}_{j=1}^{J}$, where $J < K$. The optimization objective follows the standard formulation:
\begin{equation}
\min_{{\boldsymbol{c}_j}} \sum_{k=1}^{K} \min_{j} \| \boldsymbol{e}_k - \boldsymbol{c}_j \|_2^2.
\end{equation}
Each original code $\boldsymbol{e}_k$ is then reassigned to its nearest cluster centroid $\boldsymbol{c}_j$, effectively yielding a coarser but semantically consistent representation. 

For a joint-state sequence $\boldsymbol{Q}_t$, we first obtain its latent embedding $z_t = \mathcal{E}(\boldsymbol{Q}_t)$ and the corresponding quantized codes $\boldsymbol{e}^k_t$. Each quantized code is then mapped to its nearest centroid, and the centroid index $j$ serves as the final discrete token representing the sequence $\boldsymbol{Q}_t$.
This post-hoc clustering reduces redundancy and emphasizes high-level semantic patterns shared across trajectories, rather than low-level variations, as illustrated in~\cref{fig:clustering}. Consequently, these clustered tokens yield a compact memory representation
\section{Experiments}
We conduct a series of experiments to assess the effectiveness, generality, and robustness of VQ-Memory under different learning scenarios. Specifically, we aim to answer three key questions: 
(1) How effective and general is VQ-Memory when applied to various model architectures in single-task settings? (\cref{sec:single-task})
(2) Can VQ-Memory effectively improve learning performance in multi-task settings? (\cref{sec:multi-task})
(3) How do the main hyperparameters of VQ-Memory affect its performance? (\cref{sec:ablation})

\subsection{Experimental Setup}
\begin{figure}[t]
  \centering
\includegraphics[width=0.99\linewidth]{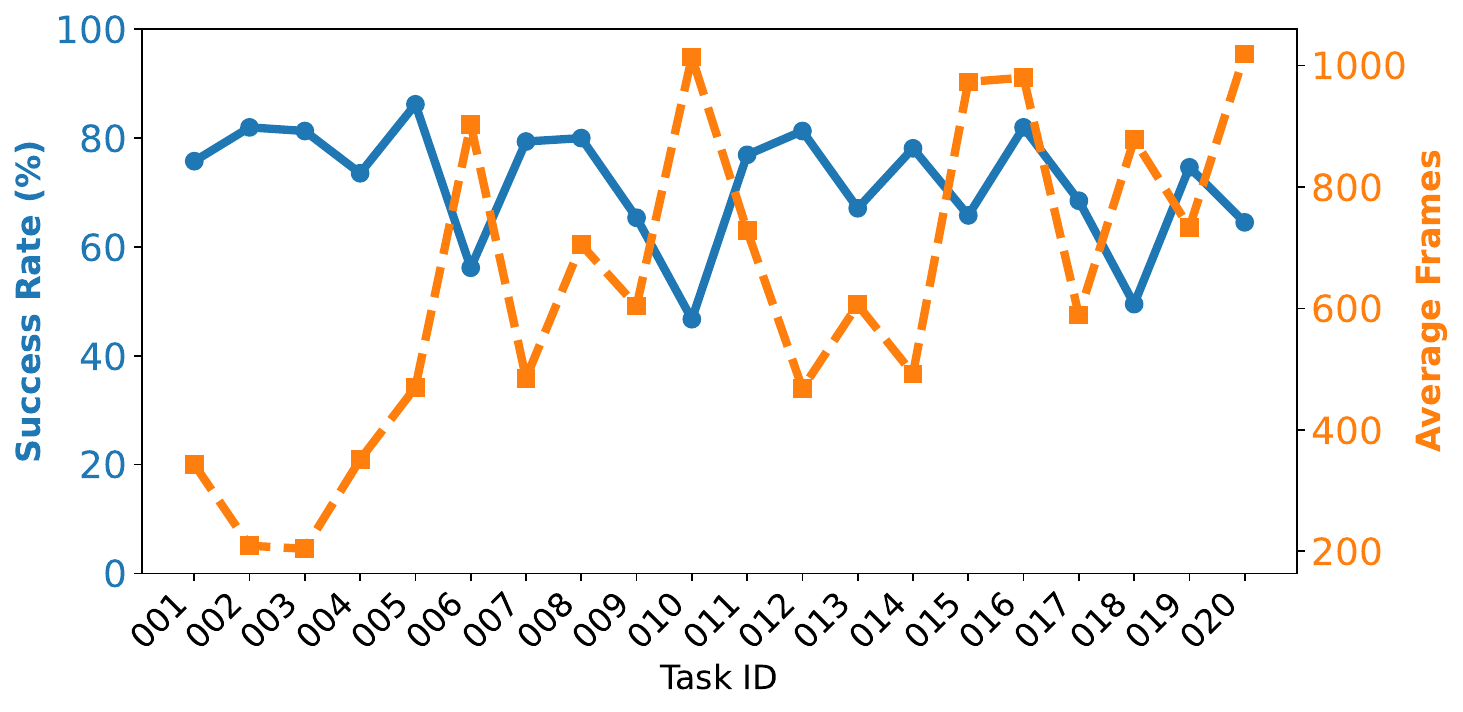}
   \caption{Statistics of Demonstration Generation in RuleSafe: Success Rate and Average Frames.}
   \label{fig:statistics}
   \vspace{-1em}
\end{figure}
\paragraph{Overview of the RuleSafe Benchmark} 
RuleSafe consists of 20 lock rules of varying complexity and 10 types of safes (See Appendix~\ref{sec:rule_and_obj} for more details). During demonstration generation, the initial location and pose of each safe are randomly varied. 
In simulation, the environment runs at 100 Hz, while data is collected at 10 Hz. 
\cref{fig:statistics} presents the success rate for generating demonstrations for each rule, along with the average demonstration length. Across all rules, the mean success rate is 71.7\% with an average trajectory length of 638 frames.

We evaluate both \textbf{single-task} and \textbf{multi-task} settings. In the single-task experiments, we focus on two representative rules shown in~\cref{fig:two_rules}. For clarity, we denote the part-phase-based rule as \textbf{\texttt{rule\_001}} and the task-phase-based rule as \textbf{\texttt{rule\_020}}. Each task/rule in the single-task setting is trained with 100 demonstrations. In the multi-task experiments, all 20 tasks are trained simultaneously with a total of 1000 trajectories (50 trajectories per task).

Performance is measured using \textbf{success rate} (SR) and \textbf{process score} (PS). The success rate represents the proportion of task executions that are fully successful, while the process score reflects the overall percentage of task steps that are correctly completed.

\paragraph{Base Models} We evaluate VQ-Memory across several representative model architectures:  
$\bullet$ \textbf{DP3}~\cite{ze20243ddiffusion} is a diffusion policy~\cite{chi2025diffusion} that takes point cloud features as input.  
$\bullet$ \textbf{RDT}~\cite{liurdt} is a large diffusion Transformer~\cite{peebles2023scalable} guided by image and language embeddings, employing an alternative injection strategy.  
$\bullet$ \textbf{CogACT}~\cite{li2024cogact} is a diffusion Transformer driven by a cognition token produced by a vision-language model (VLM).  
$\bullet$ \textbf{$\pi_0$}~\cite{black2024pi0} is a flow-matching model~\cite{lipmanflow,liu2022rectified} built upon a pre-trained VLM using a mixture-of-experts framework. 
To integrate VQ-Memory with DP3, we add a small convolutional network to map the discrete memory tokens into a latent embedding, which is then used as input to DP3.  
For other models, we treat the discrete memory tokens as special language tokens. This is achieved by mapping the VQ-Memory's dedicated vocabulary (of size $N$) to a corresponding set of IDs at the tail end of the VLM's original vocabulary. Specifically, a memory token with index $j$ is represented by the VLM's token ID $V_{\text{size}} - N + j$. The resulting sequence of memory tokens is then concatenated with the policy's existing language tokens before being processed by the Transformer encoder.

\paragraph{Implementation Details} 
All experiments are conducted on 4 NVIDIA A100 GPUs.  
For each model, we evaluate its performance with and without VQ-Memory under identical settings, using the default configurations of each method.  
DP3 takes as input 1024 points representing the point cloud and outputs 13-DoF actions. It is trained with a batch size of 256 for 200k steps.  
RDT, CogACT, and $\pi_0$ take as input two RGB images of size $224 \times 224$ from different viewpoints along with text tokens, and produce 13-DoF actions. These models are fine-tuned from their released pretrained weights for 30k steps, using a batch size of 8 for RDT and 32 for CogACT and $\pi_0$. All policies predict a 50-step action chunk.
VQ-Memory is configured with a vocabulary size of 4, clustered from an original vocabulary of 256, and a memory token length of 40.

\subsection{Effectiveness and Generality of VQ-Memory in Single-Task Setting}
\label{sec:single-task}

We first evaluate the effectiveness of VQ-Memory in providing temporal context for robust decision-making, while mitigating the overfitting issues associated with using raw joint-state sequences as memory tokens. Specifically, we assess the performance of $\pi_0$~\cite{black2024pi0} on \textbf{\texttt{rule\_001}} and \textbf{\texttt{rule\_020}} under the single-task setting, where \textbf{\texttt{rule\_001}} involves a short three-step operation and \textbf{\texttt{rule\_020}} requires completing a longer eight-step sequence.
As shown in~\cref{tab:two_rules}, the non-Markovian RuleSafe tasks pose a substantial challenge for the $\pi_0$ policy when it relies solely on current observations.
For example, when approaching a handle, the policy often struggles to determine whether it should \textit{pull} or \textit{rotate}, as the current visual observation alone does not reveal the underlying task stage.

Introducing raw historical joint states as memory tokens initially improves short-horizon planning accuracy for the \textbf{\texttt{rule\_001}} task. However, the performance becomes unstable on the long-horizon complex \textbf{\texttt{rule\_020}} task. We hypothesize that this instability arises from the continuous nature of raw joint states, which makes them highly sensitive to variations and noise. These fluctuations lead to a distribution shift between training and test-time state sequences, causing the model to overfit specific trajectories rather than learning robust and generalizable temporal patterns.

In contrast, when we replace raw joint states with VQ-Memory tokens, the model demonstrates substantial improvements across both tasks. This improvement indicates that VQ-Memory effectively mitigates overfitting while preserving essential temporal information. 

Moreover, we further examine whether the benefits of VQ-Memory generalize beyond the $\pi_0$ policy by integrating it into different model architectures, including DP3, RDT, and CogACT. As summarized in~\cref{tab:diff_model}, VQ-Memory consistently enhances the performance across all baselines on the challenging \textbf{\texttt{rule\_020}} task. These results confirm that VQ-Memory provides a versatile and architecture-agnostic mechanism for improving temporal understanding in non-Markovian manipulation tasks.

\begin{table}[t]
  \centering
  \caption{Single-task performances of different memory types. “Raw memory” refers to using raw joint-state sequences as memory tokens. We report success rate (SR) and process score (PS).}
  % \sisetup{table-format=1.2e+1}
  \small
  % \begin{adjustbox}{width=0.65\textwidth}
  \resizebox{\linewidth}{!}{
  \begin{tabular}{lccccc}
    \toprule
    %\cmidrule(r){1-4}
     \multirow{2}{*}{Method} & \multicolumn{2}{c}{\textbf{\texttt{rule\_001}}} & & \multicolumn{2}{c}{\textbf{\texttt{rule\_020}}}  \\
    \cmidrule(rl){2-3} \cmidrule(rl){5-6}
     & SR (\%) & PS (\%)& & SR (\%)& PS (\%)\\
    \midrule
    $\pi_0$ & 30.0 & 56.7 & & 0.0 & 10.6 \\
    $\pi_0$ + raw memory & 55.0 & 70.1 & & 0.0 & 16.3 \\
    % $\pi_0$ + VQ-memory & 85.0 & 93.3 & & 60.0 & 71.3 \\
    $\pi_0$ + VQ-memory & 80.0 & 89.3 & & 45.0 & 67.3 \\
    % 0.0068 before
    \bottomrule
  \end{tabular}
  }
  % \end{adjustbox}
  \label{tab:two_rules}
\end{table}

\begin{table}[h]
  \centering
  \caption{Effect of VQ-Memory across models.}
  \small
  % \begin{adjustbox}{width=0.65\textwidth}
  % \resizebox{\linewidth}{!}{
  \begin{tabular}{lcc}
    \toprule
    %\cmidrule(r){1-4}
     \multirow{2}{*}{Method} & \multicolumn{2}{c}{\textbf{\texttt{rule\_020}}}  \\ \cmidrule(rl){2-3}
     & Success Rate (\%) & Process Score (\%)\\
    \midrule
    DP3 & 5.0 & 14.1 \\
    \rowcolor{gray!20}\ + VQ-memory & 45.0 & 70.4 \\
    \midrule
    RDT & 0.0 & 13.6 \\
    \rowcolor{gray!20}\ + VQ-memory & 35.0 & 61.2 \\
    \midrule
    CogACT & 0.0 & 9.7 \\
    \rowcolor{gray!20}\ + VQ-memory & 20.0 & 50.4 \\
    \midrule
    $\pi_0$ & 0.0 & 10.6 \\
    \rowcolor{gray!20}\ + VQ-memory & 45.0 & 67.3 \\
    % 0.0068 before
    \bottomrule
  \end{tabular}
  % }
  % \end{adjustbox}
  \label{tab:diff_model}
\end{table}

\subsection{Multi-Task Learning with VQ-Memory}
\label{sec:multi-task}
After confirming the benefits of VQ-Memory in the single-task setting, we further evaluate its effectiveness under the \textbf{multi-task learning} scenario.
We train the $\pi_0$ policy with and without VQ-Memory on the full set of RuleSafe tasks, using a batch size of 64 for 60k training steps.
This configuration differs from the single-task setting, as the multi-task setting involves a larger data volume and task diversity.

\paragraph{Main Results} As shown in~\cref{tab:multi-task}, $\pi_0$ exhibits strong performance on relatively simple tasks such as \textbf{\texttt{rule\_002}} and \textbf{\texttt{rule\_003}}.
We attribute this to two factors:
(1) these tasks involve fewer operation steps, which limits temporal ambiguity; and
(2) the specific \textit{safe assets} used in these tasks contain joints whose rotation angles can be directly inferred from visual cues, enabling $\pi_0$ to infer task progress based solely on the current observation.
However, in long-horizon, $\pi_0$ struggles severely, often yielding near-zero success rates.
In contrast, integrating VQ-Memory leads to consistent and substantial improvements across nearly all tasks.
Notably, VQ-Memory improves the average success rate from 25.0\% to 56.3\% and the process score from 48.8\% to 76.5\%, indicating its strong capability to capture temporal dependencies and maintain robust performance across diverse non-Markovian manipulation tasks.

\begin{table*}[t]
    \centering
    \small
     \caption{Performance of $\pi_0$ with and without VQ-Memory on the multi-task setting. We report success rate (\%) and process score (\%).}
    \label{tab:multi-task}
    % \resizebox{\textwidth}{!}{
        \begin{tabular}{lcc|lcc|lcc}
        \toprule
             & Success & Process & \textbf{} & Success & Process  & \textbf{} & Success  & Process   \\ 
             &  Rate &  Score& \textbf{} &  Rate  &  Score   & \textbf{} &  Rate &  Score  \\ 
            \midrule
            
            \textbf{\texttt{rule\_001}} & ~ & ~ & 
            \textbf{\texttt{rule\_002}} & ~ & ~  & \textbf{\texttt{rule\_003}} & ~  & ~ \\ 
            $\pi_0$ & 35.0 & 61.7 & 
            $\pi_0$ & 80.0 & 85.0
            & $\pi_0$ & 75.0 & 87.5\\
            
            +VQ-Memory & 60.0  & 73.3 & 
            +VQ-Memory & 90.0 & 90.0 & 
            +VQ-Memory & 95.0 & 97.5 \\ 

            \midrule
            
            \textbf{\texttt{rule\_004}} & ~ & ~ & 
            \textbf{\texttt{rule\_005}} & ~ & ~  & \textbf{\texttt{rule\_006}} & ~  & ~ \\ 
            $\pi_0$ & 35.0 & 58.3 & 
            $\pi_0$ & 30.0 & 48.8
            & $\pi_0$ & 0.0 & 25.7\\
            
            +VQ-Memory & 50.0  & 66.7 & 
            +VQ-Memory & 45.0 & 56.3 & 
            +VQ-Memory & 60.0 & 83.6 \\ 

            \midrule
            
            \textbf{\texttt{rule\_007}} & ~ & ~ & 
            \textbf{\texttt{rule\_008}} & ~ & ~  & \textbf{\texttt{rule\_009}} & ~  & ~ \\ 
            $\pi_0$ & 65.0 & 87.5 & 
            $\pi_0$ & 10.0 & 48.8
            & $\pi_0$ & 25.0 & 48.0\\
            
            +VQ-Memory & 75.0  & 87.5 & 
            +VQ-Memory & 40.0 & 66.3 & 
            +VQ-Memory & 45.0 & 82.0 \\ 

            \midrule
            
            \textbf{\texttt{rule\_010}} & ~ & ~ & 
            \textbf{\texttt{rule\_011}} & ~ & ~  & \textbf{\texttt{rule\_012}} & ~  & ~ \\ 
            $\pi_0$ & 0.0 & 18.1 & 
            $\pi_0$ & 0.0 & 53.3
            & $\pi_0$ & 40.0 & 45.6\\
            
            +VQ-Memory & 20.0  & 59.4 & 
            +VQ-Memory & 60.0 & 91.7 & 
            +VQ-Memory & 45.0 & 57.5 \\ 

            \midrule
            
            \textbf{\texttt{rule\_013}} & ~ & ~ & 
            \textbf{\texttt{rule\_014}} & ~ & ~  & \textbf{\texttt{rule\_015}} & ~  & ~ \\ 
            $\pi_0$ & 15.0 & 42.0 & 
            $\pi_0$ & 55.0 & 72.5
            & $\pi_0$ & 5.0 & 35.6\\
            
            +VQ-Memory & 35.0  & 54.0 & 
            +VQ-Memory & 70.0 & 85.0 & 
            +VQ-Memory & 50.0 & 82.5 \\ 

            \midrule
            
            \textbf{\texttt{rule\_016}} & ~ & ~ & 
            \textbf{\texttt{rule\_017}} & ~ & ~  & \textbf{\texttt{rule\_018}} & ~  & ~ \\ 
            $\pi_0$ & 5.0 & 45.0 & 
            $\pi_0$ & 20.0 & 45.0
            & $\pi_0$ & 0.0 & 18.6\\
            
            +VQ-Memory & 80.0  & 96.3 & 
            +VQ-Memory & 75.0 & 90.0 & 
            +VQ-Memory & 30.0 & 70.7 \\ 

            \midrule
            
            \textbf{\texttt{rule\_019}} & ~ & ~ & 
            \textbf{\texttt{rule\_020}} & ~ & ~  & \textbf{Average} & ~  & ~ \\ 
            $\pi_0$ & 0.0 & 40.0 & 
            $\pi_0$ & 5.0 & 9.4
            & $\pi_0$ & 25.0 & 48.8\\
            
            +VQ-Memory & 65.0  & 83.3 & 
            +VQ-Memory & 35.0 & 55.6 & 
            +VQ-Memory & 56.3{\scriptsize \textcolor{ForestGreen}{(+31.3)}} & 76.5{\scriptsize \textcolor{ForestGreen}{(+27.7)}} \\ 
            
            \bottomrule
        \end{tabular}    
    % \vspace{-2em}
\end{table*}

\paragraph{Possible Causes of Remaining Failures}
Although VQ-Memory substantially improves temporal reasoning and task planning, the overall success rate (56.3\%) indicates that precise action execution remains a key bottleneck. We observe that the policy often generates correct high-level plans—accurately reasoning about unlocking and opening sequences—but occasionally fails to execute individual operations with sufficient precision, resulting in finger misalignment, slipping, or incomplete rotations. This limitation is primarily due to the limited number of demonstrations for each safe instance. While the dataset includes multiple safes with diverse geometries and joint tolerances, the per-configuration data is insufficient for the model to fully adapt its control policy to such heterogeneous dynamics. Expanding the data per safe or applying targeted fine-tuning could further improve low-level manipulation accuracy.

\begin{table}[t]

  \centering
  \caption{Ablation on number of clusters and memory length. We report the success rate (\%) and process score (\%) on \textbf{\texttt{rule\_020}}.}
  % \sisetup{table-format=1.2e+1}
  \small
  % \begin{adjustbox}{width=0.65\textwidth}
  \begin{tabular}{c|c|cc}
    \toprule
    %\cmidrule(r){1-4}
    & Variant & Success Rate & Process Score \\
    \midrule
     \multirow{4}{*}{\shortstack{Number of\\Clusters}} & 256 & 20.0 & 47.5 \\
     & 32 & 35.0 & 57.5 \\
     & \cellcolor{gray!20}4 & \cellcolor{gray!20}\textbf{45.0} & \cellcolor{gray!20}\textbf{67.3} \\
     % & 4 & 50.0 & 67.3 \\
     & 2 & 30.0 & 54.4 \\
     \midrule
     \multirow{3}{*}{\shortstack{Memory \\Length}} 
     & 20 & 25.0 & 53.8 \\
     & \cellcolor{gray!20}40 & \cellcolor{gray!20}\textbf{45.0} & \cellcolor{gray!20}\textbf{67.3} \\
     & 60 & 40.0 & 65.4 \\
    \bottomrule
  \end{tabular}
  % \end{adjustbox}
  \label{tab:ablation}
\end{table}

\subsection{Ablation Studies}
\label{sec:ablation}
To understand the impact of key hyperparameters in VQ-Memory, we perform ablation experiments on \textbf{\texttt{rule\_020}}.

\paragraph{Number of Clusters} As shown in Table~\ref{tab:ablation}, clustering plays a key role in making the memory tokens more compact and reducing overfitting. 
% When no clustering is applied (256-way vocabulary), the performance is limited due to redundant and noisy memory representations. 
When no clustering is applied and the memory uses a vocabulary of 256 codes, performance is limited, achieving only a 20\% success rate due to redundant and noisy memory representations. Reducing the number of clusters improves generalization, with the best result achieved at 4 clusters, where the success rate rises to 45\%.
Further decreasing the cluster count to 2 slightly degrades performance, 
% implying that too coarse a clustering hampers the model’s ability to capture stage-level variations within a task.
suggesting that overly coarse clustering merges distinct stage-level patterns, hindering the model’s capacity to differentiate task progression.

\paragraph{Memory Length} We further examine the effect of memory length, varying it from 20 to 60 tokens. A shorter memory (20) fails to capture long-term dependencies, resulting in a low success rate of 25.0\%. Increasing the length from 20 to 40 considerably improves performance, while further increasing it to 60 brings negligible benefits, indicating that the improvement has reached saturation. Hence, we set the memory length to 40 in our experiments for an optimal trade-off between efficiency and temporal coverage.

\section{Conclusion}
\label{sec:conclusion}

We presented \textbf{RuleSafe} and \textbf{VQ-Memory}, two complementary contributions for long-horizon articulated manipulation. \textbf{RuleSafe} provides a scalable benchmark with non-Markovian, multi-stage tasks for evaluating temporal reasoning and planning.
To address the temporal challenges posed by these tasks, we proposed \textbf{VQ-Memory}, a compact and structured memory module that encodes past joint states into discrete latent tokens via vector quantization.
% , filtering low-level noise while preserving stage-level context. 
This design filters out low-level noise while preserving stage-level context, enabling efficient and robust long-horizon control.
Future work includes scaling RuleSafe with more rules and objects, as well as expanding VQ-Memory to encode additional sensory inputs, such as learning adaptive features~\cite{wang2025emulating} or latent action embeddings~\cite{bu2025univla} from video.

{
    \small
    \bibliographystyle{ieeenat_fullname}
    \bibliography{main}
}

\clearpage
\setcounter{page}{1}
\maketitlesupplementary

% \section{Rationale}
% \label{sec:rationale}
% % 
% Having the supplementary compiled together with the main paper means that:
% % 
% \begin{itemize}
% \item The supplementary can back-reference sections of the main paper, for example, we can refer to \cref{sec:intro};
% \item The main paper can forward reference sub-sections within the supplementary explicitly (e.g. referring to a particular experiment); 
% \item When submitted to arXiv, the supplementary will already included at the end of the paper.
% \end{itemize}
% % 
% To split the supplementary pages from the main paper, you can use \href{https://support.apple.com/en-ca/guide/preview/prvw11793/mac#:~:text=Delete%20a%20page%20from%20a,or%20choose%20Edit%20%3E%20Delete).}{Preview (on macOS)}, \href{https://www.adobe.com/acrobat/how-to/delete-pages-from-pdf.html#:~:text=Choose%20%E2%80%9CTools%E2%80%9D%20%3E%20%E2%80%9COrganize,or%20pages%20from%20the%20file.}{Adobe Acrobat} (on all OSs), as well as \href{https://superuser.com/questions/517986/is-it-possible-to-delete-some-pages-of-a-pdf-document}{command line tools}.

\appendix
\section{RuleSafe Benchmark Details}
\label{sec:rule_and_obj}
% \begin{description}[leftmargin=2cm,style=nextline]
%     \item[rule\_001] The safe door remains locked unless both the knob and handle are open.
%     \item[rule\_002] The safe door remains locked unless the knob is open.
%     \item[rule\_003] The safe door remains locked unless the handle is open.
%     % ...继续填写
% \end{description}

\paragraph{Rule Description}\cref{tab:20-rules} lists the 20 rule-based manipulation tasks used in our experiments. 
Each task defines a unique temporal or logical dependency between the knob, handle, and safe door. 
These rules vary in structure—from simple joint-state conditions (e.g., both components open) to multi-step temporal dependencies requiring password-like input sequences. 
Such diversity enables systematic evaluation of reasoning and planning.

\begin{figure}[h]
  \centering
\includegraphics[width=0.99\linewidth]{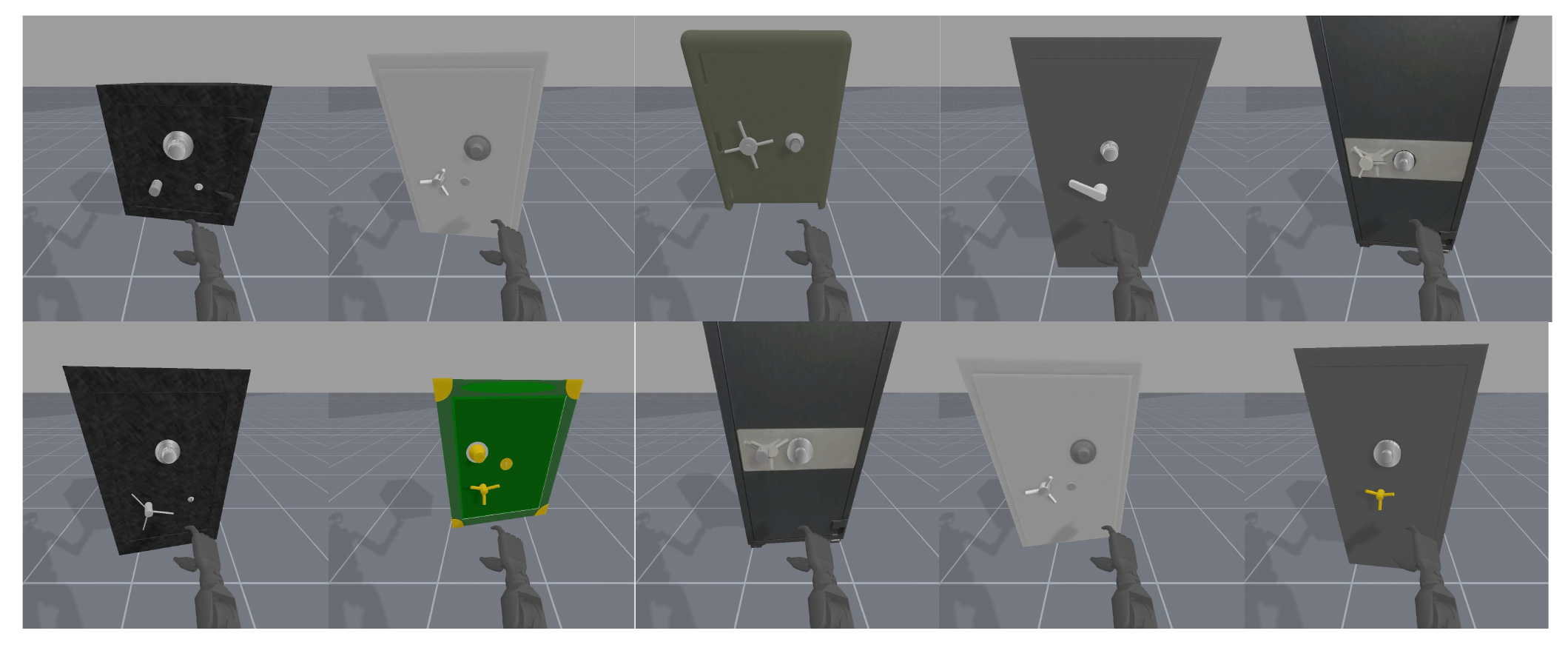}
   \caption{List of 10 object instances used in RuleSafe.}
   \label{fig:obj}
\end{figure}

\paragraph{Object Set}
\cref{fig:obj} summarizes the ten objects used in the benchmark. 
Objects vary in geometry, articulation type, and visual appearance, introducing additional diversity. 
Each object instance preserves the same functional semantics (knob–handle–door) but differs in physical configuration, enabling systematic evaluation of perception generalization.

% \paragraph{Spatial Annotations}
% For each hand atomic operation, we annotate key points and axes that represent the contact locations and motion directions of the fingers. Similarly, for each manipulable joint of an articulated asset, we annotate key points and axes that indicate its operational interface. 
\begin{figure}[h]
  \centering
\includegraphics[width=0.8\linewidth]{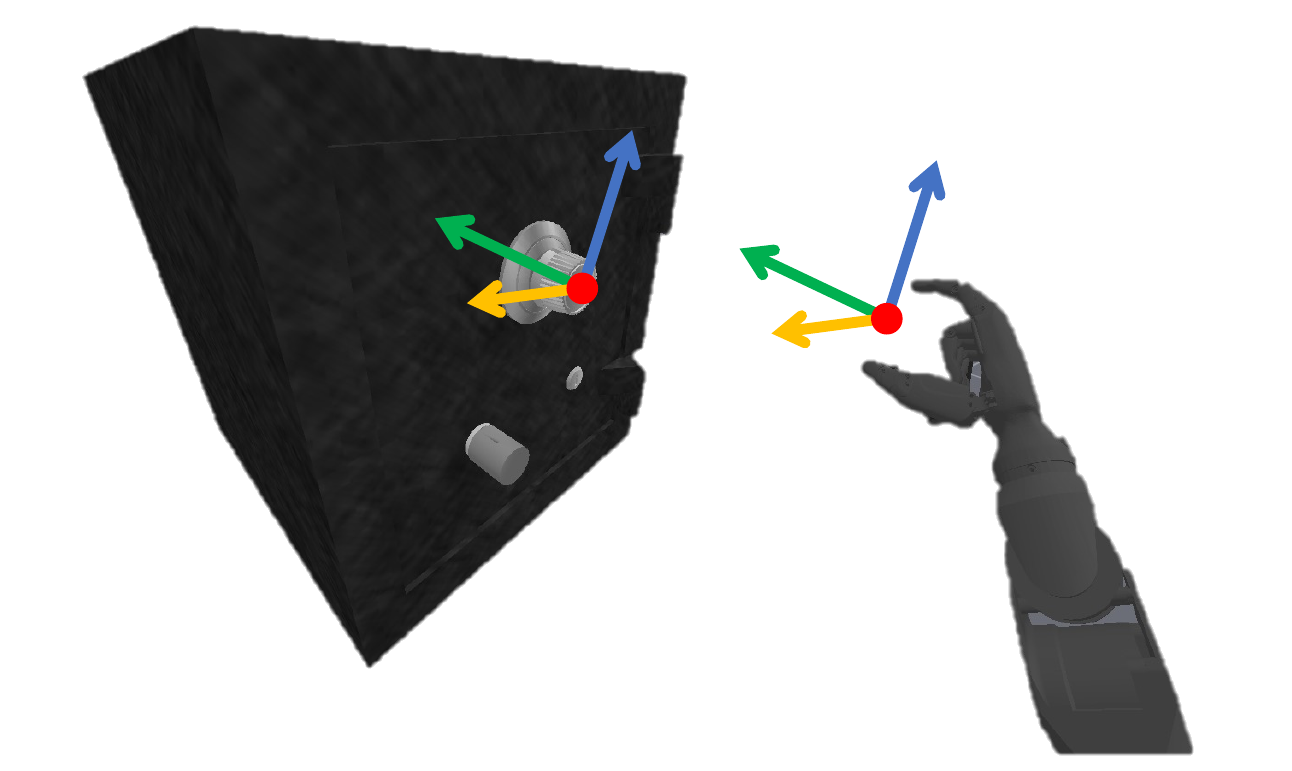}
   \caption{Annotation example. For hands, we annotate contact points and motion directions. For articulated components such as knobs, we label key points and axes representing their operational interfaces.}
   \label{fig:annotation}
\end{figure}

\onecolumn
\begin{longtable}{p{0.12\textwidth}p{0.6\textwidth}}
\caption{Descriptions of all 20 rules in the RuleSafe benchmark.} \\
\label{tab:20-rules}\\
\toprule
\textbf{Rule ID} & \textbf{Description} \\
\midrule
\endfirsthead
\toprule
\textbf{Rule ID} & \textbf{Description} \\
\midrule
\endhead
\bottomrule
\endfoot

\texttt{rule\_001} & The safe door remains locked unless both the knob and handle are open. \\
\texttt{rule\_002} & The safe door remains locked unless the knob is open. \\
\texttt{rule\_003} & The safe door remains locked unless the handle is open. \\
\texttt{rule\_004} & The safe door remains locked unless both the knob and handle are open. Additionally, the knob cannot be opened unless the handle is open. \\
\texttt{rule\_005} & The safe door remains locked unless the password '001' is entered. Changing the knob's state from open to closed (or vice versa) inputs a '0', while changing the handle's state inputs a '1'. \\
\texttt{rule\_006} & The safe door remains locked unless the password '010' is entered. When handle is open, changing the knob's state from open to closed (or vice versa) inputs a '0'; otherwise it inputs a '1'. \\
\texttt{rule\_007} & The safe door unlocks when the knob and handle are in opposite states (one open, one closed). Both must have been toggled at least once. \\
\texttt{rule\_008} & The safe door unlocks only if the knob is open and the handle has been toggled an even number of times greater than one. \\
\texttt{rule\_009} & The safe door unlocks after receiving binary input '1001'. Knob state change = '1', handle state change = '0'. Input buffer holds last 4 digits. \\
\texttt{rule\_010} & The safe door unlocks when the password '0110' is entered. Handle state change while knob is open inputs '0', handle change while knob closed inputs '1'. \\
\texttt{rule\_011} & The safe door unlocks after the password '1010' is entered. Knob state open→closed = '1', closed→open = '0'. Handle changes ignored. \\
\texttt{rule\_012} & The safe door unlocks after knob and handle have been in all possible state combinations (open/open, open/closed, closed/open, closed/closed). \\
\texttt{rule\_013} & The safe door unlocks when the handle is open and the knob has changed state a non-zero number of times that is divisible by 3. \\
\texttt{rule\_014} & The safe door remains locked unless the knob is closed and the handle is open. Additionally, the handle cannot be opened unless the knob is open. \\
\texttt{rule\_015} & The safe door unlocks when knob open count equals handle close count (minimum 2 each). Counts reset if either reaches 3. \\
\texttt{rule\_016} & The safe door unlocks after the password '1110' is entered. Closing the knob inputs '1', opening the handle inputs '0'. Input buffer holds last 4 digits. \\
\texttt{rule\_017} & The safe door unlocks when the product of the knob and handle toggle counts equals 4. A toggle is defined as any state change. \\
\texttt{rule\_018} & The safe door unlocks when the password '123' is entered. A knob state change increments a counter, a handle state change appends the counter value to the buffer and resets the counter. The counter starts at 1. The buffer holds the last 3 digits. \\
\texttt{rule\_019} & The safe door unlocks after entering the password '01' via knob interactions (0 = knob closes, 1 = knob opens). Knob events are recorded only when the handle is open. The door unlocks only when the handle is closed after the correct password has been entered. \\
\texttt{rule\_020} & The safe door unlocks after entering the password '11' via knob interactions (0 = knob closes, 1 = knob opens). Knob events are recorded only when the handle is open. The door unlocks only when the handle is closed after the correct password has been entered. \\
\end{longtable}

\begin{lstlisting}[style=pythoncode, caption={Lock checking logic for \texttt{rule\_001} and  \texttt{rule\_020} generated by LLM in Python code form.}, label={lst:check_lock}]
# rule_001
# Rule Description: The safe door unlocks after both the knob and handle are open. 
def check_lock(self):
    # Query the open status of each component of the safe
    is_knob_open = self.is_open(obj_name='safe', link_name='knob')
    is_handle_open = self.is_open(obj_name='safe', link_name='handle')
    is_door_open = self.is_open(obj_name='safe', link_name='door')
    
    # If either the knob or handle is not open,
    # the door remains locked; otherwise, it can be unlocked
    if (not is_knob_open or not is_handle_open):
        self.set_lock(obj_name='safe', link_name='door', lock=True)
    else:
        self.set_lock(obj_name='safe', link_name='door', lock=False)
        
# rule_020
# Rule Description: The safe door unlocks after entering the password "11" via knob interactions 
# (0 = knob closes, 1 = knob opens). Knob events are recorded only when the handle is open. 
# The door unlocks only when the handle is closed after the correct password has been entered.
def check_lock(self):
 # Query the open states of knob, handle, and door
    knob = self.is_open(obj_name='safe', link_name='knob')
    handle = self.is_open(obj_name='safe', link_name='handle')
    door = self.is_open(obj_name='safe', link_name='door')

    # Initialize task state if not already created
    if not hasattr(self, '_mech_state'):
        self._mech_state = {}

    state = self._mech_state
    state.setdefault('last_knob', knob)
    state.setdefault('input_buffer', [])

    # Allow password input only when the handle is open
    if handle:
        # Record knob rotation events ("1" for open, "0" for close)
        if knob != state['last_knob']:
            event = '1' if knob else '0'
            state['input_buffer'].append(event)
            state['input_buffer'] = state['input_buffer'][-2:]  # Keep last two events only

    # Unlock the door if correct input ("11") is entered and handle is closed,
    # or if the door is already open
    if (''.join(state['input_buffer']) == '11' and not handle) or door:
        self.set_lock(obj_name='safe', link_name='door', lock=False)
    else:
        self.set_lock(obj_name='safe', link_name='door', lock=True)

    # Update the last knob state
    state['last_knob'] = knob
    
\end{lstlisting}

\begin{lstlisting}[style=pythoncode, caption={Task decomposition for \texttt{rule\_001} generated by LLM in Python code form, which serves as the foundation of the demonstration generation pipeline.}, label={lst:llm-decomp}]
# Step 1: rotate knob counterclockwise
### Hand operation: pre pinch
planner.hand_pre_pinch("right")
### Arm movement: move to knob
constraint_left = planner.generate_constraints(obj_name="safe", target_link="knob")
_, target_effector_pose = planner.generate_end_effector_pose(constraint_left)
planner.move_to_pose(target_effector_pose)
### Hand operation: pinch knob
planner.hand_pinch("right", pinch_object="safe")
### Hand operation: rotate
planner.rotate_hand(hand="right", angle=-60)

# Step 2: rotate handle counterclockwise
### Hand operation: release fingers
planner.hand_pre_pinch("right")
### Arm movement: move to handle
constraint_left = planner.generate_constraints(obj_name="safe", target_link="handle")
_, target_effector_pose = planner.generate_end_effector_pose(constraint_left, hand_name="right")
planner.move_to_pose(target_effector_pose)
### Hand operation: pinch handle
planner.hand_pinch("right", pinch_object="safe")
### Hand operation: rotate
planner.rotate_hand(hand="right", angle=-60)

# Step 3: pull handle
### Arm movement: move to the target pose corresponding to the door state with openness = 1
constraint_left = planner.generate_constraints(obj_name="safe", target_link="handle", state_link="door", openness=1)
_, target_effector_pose = planner.generate_end_effector_pose(constraint_left)
planner.move_to_pose(target_effector_pose)
\end{lstlisting}
\twocolumn

\end{document}